\documentclass[conference]{IEEEtran}
\IEEEoverridecommandlockouts

\usepackage{cite}
\usepackage{amsmath,amssymb,amsfonts}
\usepackage{algorithmic}
\usepackage{graphicx}
\usepackage{textcomp}
\usepackage{xcolor}
\def\BibTeX{{\rm B\kern-.05em{\sc i\kern-.025em b}\kern-.08em
    T\kern-.1667em\lower.7ex\hbox{E}\kern-.125emX}}
\def\ie{\emph{i.e.}}
\def\etc{\emph{etc.}}
\begin{document}

\title{Learning Effective NeRFs and SDFs Representations with 3D GANs 
for Object Generation}

\author{\IEEEauthorblockN{1\textsuperscript{st} Zheyuan Yang *}
\IEEEauthorblockA{\textit{Department of Electrical and Computer Engineering} \\
\textit{University of Toronto}\\
Toronto, Canada \\
andrewzheyuan.yang@mail.utoronto.ca}
\thanks{* Equal Contribution. Work done during an internship at Noah Ark's Lab.}
\and
\IEEEauthorblockN{2\textsuperscript{nd} Yibo Liu *}
\IEEEauthorblockA{\textit{Department of Earth and Space Science} \\
\textit{York University}\\
Toronto, Canada \\
buaayorklau@gmail.com}
\and
\IEEEauthorblockN{3\textsuperscript{rd} Guile Wu}
\IEEEauthorblockA{\textit{Huawei Noah Ark's Lab} \\
Toronto, Canada \\
guile.wu@huawei.com}
\and
\IEEEauthorblockN{4\textsuperscript{th} Tongtong Cao}
\IEEEauthorblockA{\textit{Huawei Noah Ark's Lab} \\
Toronto, Canada \\
caotongtong@huawei.com}
\and
\IEEEauthorblockN{5\textsuperscript{th} Yuan Ren}
\IEEEauthorblockA{\textit{Huawei Noah Ark's Lab} \\
Toronto, Canada \\
yuan.ren3@huawei.com}
\and
\IEEEauthorblockN{6\textsuperscript{th} Yang Liu}
\IEEEauthorblockA{\textit{Huawei Noah Ark's Lab} \\
Toronto, Canada \\
yang.liu9@huawei.com}
\and
\IEEEauthorblockN{7\textsuperscript{th} Bingbing Liu}
\IEEEauthorblockA{\textit{Huawei Noah Ark's Lab} \\
Toronto, Canada \\
liu.bingbing@huawei.com}
}

\maketitle

\begin{abstract}
We present a solution for 3D object generation of ICCV 2023 OmniObject3D Challenge.
In recent years, 3D object generation has made great process and achieved promising results,
but it remains a challenging task due to the difficulty of generating complex, textured and high-fidelity results.
To resolve this problem, we study learning effective NeRFs and SDFs representations
with 3D Generative Adversarial Networks (GANs) for 3D object generation.
Specifically, inspired by recent works,
we use the efficient geometry-aware 3D GANs as the backbone incorporating with label embedding and color mapping,
which enables to train the model on different taxonomies simultaneously.
Then, through a decoder, we aggregate the resulting features to generate Neural Radiance Fields (NeRFs) based representations
for rendering high-fidelity synthetic images.
Meanwhile, we optimize Signed Distance Functions (SDFs) to effectively represent objects with 3D meshes.
Besides, we observe that this model can be effectively trained with only a few images of each object from a variety
of classes, instead of using a great number of images per object or training one model per class.
With this pipeline, we can optimize an effective model for 3D object generation.
This solution is among the top 3 in the ICCV 2023 OmniObject3D Challenge.
\end{abstract}


\section{Introduction}
\label{sec:intro}
3D object generation aims at generating meaningful 3D surfaces and synthesis images of 3D objects given random inputs \cite{omni,jiang2023sdf}.
Inspired by the great success of 2D object generation \cite{karras2019style,esser2021taming},
there have been many studies \cite{lunz2020inverse,gadelha20173d} extending 2D methods to 3D generation.
Recent 3D generation approaches focus more on effective 3D representation learning \cite{nfi,mo2019structurenet} and
advanced generation strategy \cite{stylegan,eg3d,rodin}.
Despite the great progress and promising results in recent years,
3D object generation is still a challenging task due to the difficulty of generating complex, textured and high-fidelity results.

To devise a solution for 3D object generation in the ICCV 2023 OmniObject3D Challenge,
we simultaneously consider effective 3D representation learning and advanced generation strategy.
In terms of 3D representation learning, textured mesh \cite{txtmesh} and Neural Radiance Fields (NeRFs) \cite{nerf}
embrace the great potential to effectively encode 3D object features.
In particular, textured mesh representations can support individual texture map generation,
which allows us to easily replace surface textures,
while NeRFs-based representations excel at rendering high-fidelity images.
Furthermore, when considering effective shape representation, there are a variety of options,
ranging from explicit representations like point-based \cite{pointnet}, mesh-based \cite{mesh1} and voxel-based \cite{voxel1} approaches
to implicit ones such as the Signed Distance Functions (SDFs) \cite{deepsdf,mvdeepsdf} and occupancy function \cite{occ}.
In light of these cutting-edge techniques,
we mainly focus on exploring effective NeRFs representations for 3D object generation
while still optimizing SDFs representations to effectively represent objects with 3D meshes.
\begin{figure*}[t]
	\centering
	\includegraphics[width=17.5cm]{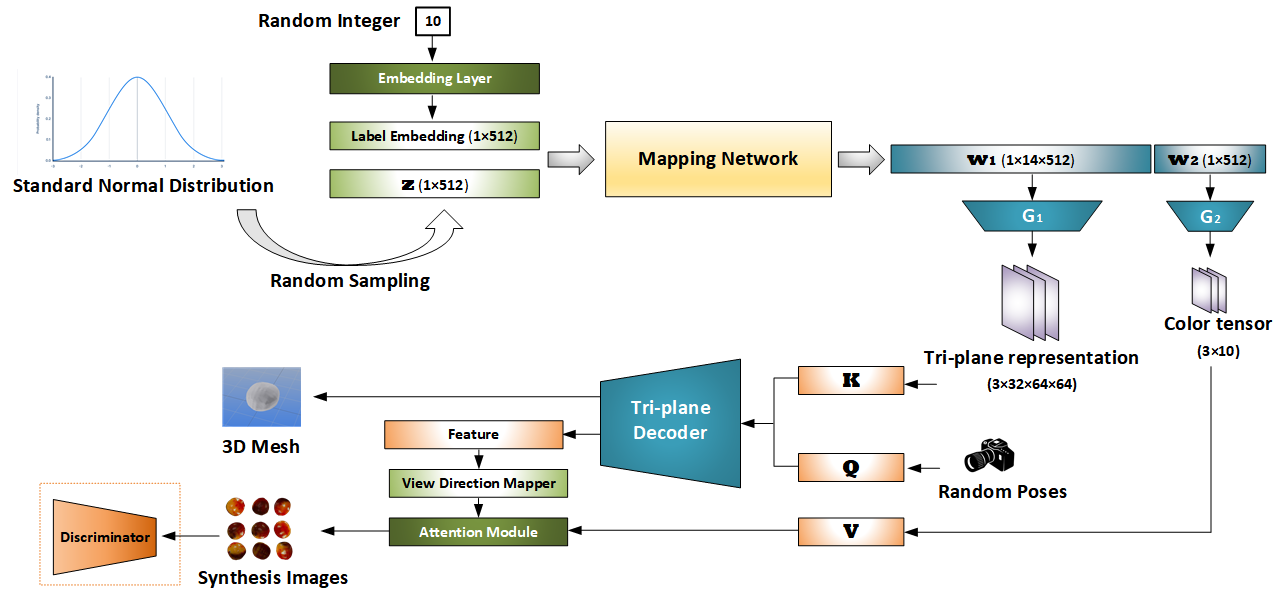}
	\caption{An overview of the framework of our solution for 3D object generation.}
	\label{pipeline}
\end{figure*}
Moving on to advanced generation strategy, there are usually different options,
such as latent generation \cite{rodin} and direct generation \cite{diffrf}.
In particular, latent generation methods aim to generate the representation of the desired output in the latent space
and decode the latent code back to the 3D space,
while direct generation focus on directly generating the desired 3D representation without utilizing the latent space.
Besides, from the perspective of generation process, generation strategies can be categorized into Generative Adversarial Networks
(GANs) \cite{stylegan,eg3d}, Variational AutoEncoder (VAE) \cite{vae}, flow-based \cite{flow}, Diffusion Models (DMs) \cite{rodin}, \etc.
In our solution, we employ the efficient geometry-aware 3D GANs \cite{eg3d} as the backbone for 3D object generation. The overall framework of our solution is depicted in Fig.~\ref{pipeline}.
Specifically, we aim to learn effective NeRFs and SDFs representations with
3D Generative Adversarial Networks (GANs) for 3D object generation.
Inspired by the recent success of efficient geometry-aware 3D GANs, \ie, EG3D \cite{eg3d,nfi},
we employ it as the backbone and incorporate label embedding and color mapping \cite{nfi}
to enable model training on different taxonomies simultaneously.
Next, we generate NeRFs-based representations for rendering high-fidelity synthetic images
and SDFs to effectively represent objects with 3D meshes through a decoder.
In addition, we empirically find that we can effectively train this model with only a few images per object from various classes,
rather than training with a great number of images per object or optimizing one model per class.
With this pipeline, we can optimize an effective model for 3D object generation.
This solution is one of the final top-3-place solutions in the ICCV 2023 OmniObject3D Challenge.

\begin{figure*}[t]
    \centering
    \includegraphics[width=17.5cm]{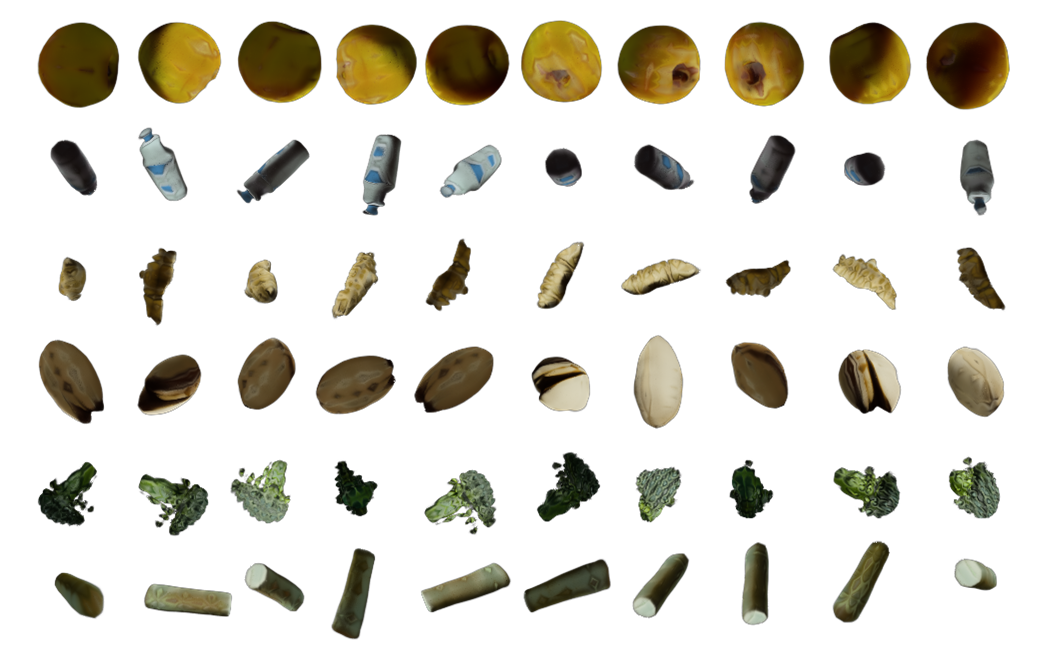}
    \caption{Images rendered by our solution with label embedding.}
    \label{result}
\end{figure*}
\section{Related Work} \label{related}
3D assets \cite{obja,omni,shapenet} are crucial for robotics and computer vision \cite{ren} because they provide accurate spatial representations and realistic environments essential for training, simulation, and object recognition. Unfortunately, the collection of 3D assets is time-consuming and labor-intensive \cite{lpr,omni}. For example, when constructing the large-scale dataset, OmniObject3D \cite{omni}, the objects are placed in a calibrated environment \cite{omni,lpr} built using fiducial markers \cite{ap3,intensity,improvements}, and the 3D scanners sample the objects from different perspectives. These fiducial markers, though low-cost \cite{shuo1, liao, shuo2, navigation,shuo3}, require additional labor to deploy in the environment. To acquire high-quality scanning results and avoid motion blur caused by aggressive sensor movement \cite{deblur}, it takes 15 minutes to one hour to finish the sampling of an object, depending on the geometric complexity of the object. Generating 3D assets \cite{stylegan,eg3d,nfi,vqadiff} is of great significance because it allows us to learn from limited real-world data with constrained renderings, enabling the creation of more diverse, photorealistic, and scalable 3D content for applications in robotics, computer vision, and beyond.
\par
The field of 3D reconstruction and generation has seen significant advancements in recent years, driven by the development of neural rendering techniques. Neural Radiance Fields (NeRF) \cite{nerf} pioneered the use of coordinate-based neural networks to represent 3D scenes as continuous volumetric functions, enabling high-quality novel view synthesis. However, NeRF's reliance on dense input views and its computationally intensive training process limit its scalability. To address these issues, NeuS \cite{neus} introduced a signed distance function (SDF)-based \cite{deepsdf,mvdeepsdf} representation, which improves surface reconstruction accuracy by explicitly modeling the geometry of objects. This approach achieves better performance in scenarios with sparse input views and complex topologies.
\par
Building on these advancements, recent works have explored efficient and compact representations for 3D data. Among these, the triplane latent representation proposed in \cite{eg3d} is a milestone for 3D generation, where 3D shapes are encoded into three orthogonal feature planes. This representation strikes a balance between expressiveness and computational efficiency, enabling faster training and inference while maintaining high fidelity.
\par
Leveraging the triplane latent representation, researchers have integrated it with generative models to achieve promising results in 3D synthesis. GAN-based methods \cite{nfi,gan2} utilize adversarial training to generate realistic 3D shapes by learning the distribution of triplane codes from large-scale datasets. On the other hand, DMs \cite{triplanediff,rodin} have recently gained traction for their ability to generate highly detailed and coherent 3D structures. By iteratively denoising triplane codes, diffusion-based approaches achieve superior control over the generation process, enabling applications such as text/image-to-3D. 

\begin{figure*}[t]
    \centering
    \includegraphics[width=17.5cm]{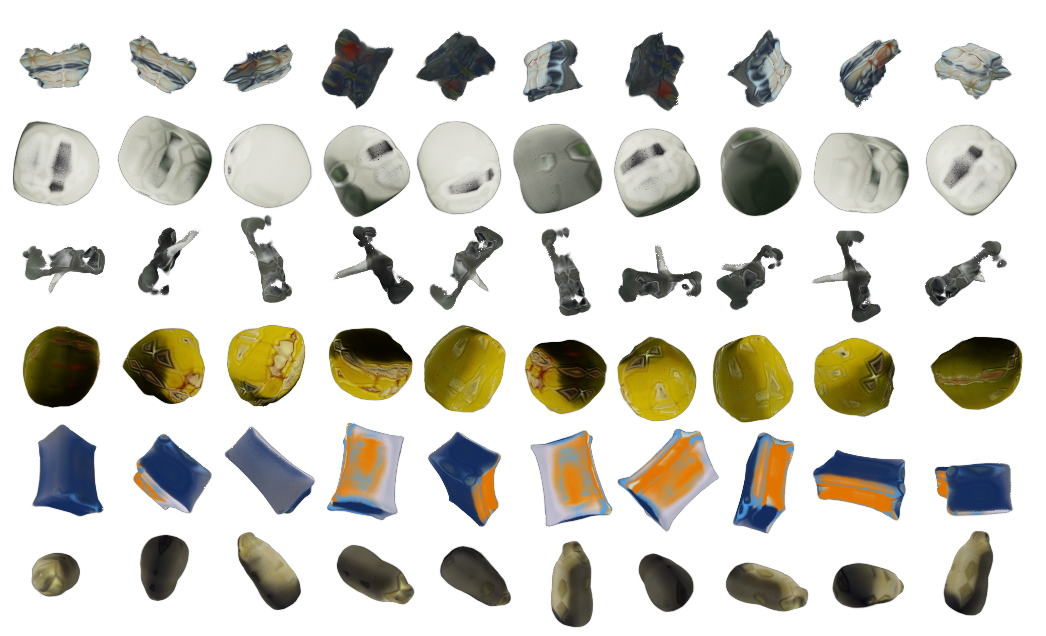}
    \caption{Images rendered by our solution without label embedding.}
    \label{result_wo_label}
\end{figure*}

\section{Method}
\subsection{Motivation of Using GANs instead of DMs} \label{mov}
Although diffusion models have recently gained an increasing popularity in 3D object generation,
most existing works \cite{diffrf,rodin} have only been verified to work under specific category conditions
(e.g., chairs \cite{diffrf} or avatars \cite{rodin}).
Besides, since the OmniObject3D dataset \cite{omni} is a collection of objects from hundreds of taxonomies,
following the pipeline of existing DMs would necessitate the use of hundreds of pre-trained models, which is impractical.
Hence, we employ a method that enables simultaneous training on all the taxonomies in the OmniObject3D dataset.
Moreover, given the success of StyleGAN2 \cite{stylegan} in representing NeRFs in latent space,
we have chosen latent generation for our pipeline.

\subsection{Solution Pipeline}
The framework of our solution is depicted in Fig.~\ref{pipeline}.
Overall, our solution adopts a GAN-based generation process.
Specifically, with EG3D \cite{eg3d} as the backbone,
firstly, we generate a one-dimensional latent code, $\mathbf{z}$ ($1\times512$),
which is obtained by randomly sampling 512 values from a standard normal distribution.
Meanwhile, we randomly generate an integer, indicating the object label, within the range of the total category number of OmniObject3D.
The integer is then transformed into a label embedding by an embedding layer.
This enables the model to be trained on different taxonomies simultaneously.
Next, we forward the label embedding and latent code $\mathbf{z}$ into a mapping network
and generate an intermediate latent code $\mathbf{w}$ ($1\times15\times512$) with promoted dimensions.
Here, the first 14 dimensions of $\mathbf{w}$ ($\mathbf{w}_{1}$ in the figure) are used as the input to the object generator
($\mathbf{G}_{1}$ in the figure) while the last dimension of $\mathbf{w}$ ($\mathbf{w}_{2}$ in the figure)
are employed as the input to the color generator ($\mathbf{G}_{2}$ in the figure).
$\mathbf{G}_{1}$ and $\mathbf{G}_{2}$ generate the tri-plane representation ($3\times32\times64\times64$)
and the color tensor ($3\times10$), respectively. 
Following \cite{nfi}, we use the color generator which allows convenient texture conversion.
After that, we determine queries, $\mathbf{Q}$, based on the randomly generated camera pose.
The tri-plane representation is then taken as keys, $\mathbf{K}$, and transmitted into the tri-plane decoder together with $\mathbf{Q}$.
Then, we use a decoder to generate the SDF of the object, which can be sampled to create a 3D mesh.
This decoder also produces a feature as the input to a view direction mapper,
from which the output is multiplied by the color tensor, $\mathbf{V}$ (the values), in the attention module.
Finally, the attention module produces the 2D synthesis image
and a discriminator \cite{eg3d} which only functions during training is used for adversarial training.

\subsection{Model Training}
Since we are simultaneously optimizing NeRFs and SDFs representations,
we ensure the generator can initially generate a unit sphere by initializing our model with SDF-pretraining as \cite{nfi}.
Then, the generator is optimized using the Adam optimizer with a learning rate of 0.0025,
while the discriminator \cite{eg3d} uses a learning rate of 0.002.
The batch size is set to 32 and the model is trained for 300,000 iterations.
Additionally, we also use adaptive discriminator augmentation \cite{ada} to improve model generalization.
Overall, the model training objective is:
\begin{equation}
\alpha_{0}\mathcal{L}_{path}+\alpha_{1}\mathcal{L}_{e}+\alpha_{2}\mathcal{L}_{v}+\alpha_{3}\mathcal{L}_{sdf},
\end{equation}
where $\mathcal{L}_{path}$ denotes the Path length regulation loss \cite{stylegan},
$\mathcal{L}_{e}$ and $\mathcal{L}_{v}$ denote the entropy loss and total variation loss \cite{stylegan}, respectively,
$\mathcal{L}_{sdf}$ represents the SDF eikonal loss \cite{nfi},
and $\alpha_{i}$ are weighting parameters.
We set $\alpha_{0}{=}2$, $\alpha_{1}{=}0.05$, $\alpha_{2}{=}0.5$ and $\alpha_{3}{=}0.1$.


\section{Experiments}
\subsection{Dataset}
We train our model from scratch for 3D object generation on the training dataset provided by
the ICCV 2023 OmniObject3D Challenge.
The complete OmniObject3D dataset \cite{omni} contains 216 classes from a wide range of daily categories
and approximately 6k objects in total where each object has 100 images in different views. 
\subsection{Efficient Model Training}
Instead of training one model for each class, we train a single model with various types of objects.
The model takes each image along with its camera pose, focal length, and object type.
Besides, we empirically find that we can effectively train this model with only a few images per object from various classes,
so we select the first 8 images of each object for model training.
This improves the model training efficiency.

\begin{table}[t]
    \centering
    \caption{Improvement in terms of FID of our solution after using the additional label embedding.}
    \label{table}
    \begin{tabular}{c|c}
        \hline \hline
        Method & FID \\
        \hline
        NFI (CVPR'23) \cite{nfi} & 62.426 \\
        Ours w/o label embedding & 62.335 \\
        Ours w/ label embedding & \textbf{57.624} \\
        \hline \hline
    \end{tabular}
   
\end{table}
\subsection{Results}
Fig.~\ref{result} shows some visualization results of synthesis images generated by our solution on OmniObject3D.
It can be seen that our approach is capable of generating high-fidelity images from different views for 3D object generation.
Besides, Fig.~\ref{result_wo_label} shows results of our solution without the additional label embedding.
By comparing Fig.~\ref{result} with Fig.~\ref{result_wo_label} and the results in Table \ref{table}, we can see that without the additional label embedding, the quality of the synthesized images is significantly worse than that of the solution with label embedding. This verifies that the additional label embedding is beneficial for improving model generalization and performance. Furthermore, we compare our method with a state-of-the-art approach proposed in CVPR 2023, nerf-from-image (NFI) \cite{nfi}, as shown in Table \ref{table}, in terms of FID.
\subsection{Discussion}
Our solution is one of the follow-up works of EG3D \cite{eg3d} that leverages 3D GANs and triplane latent representation. On the other hand, the combination of DMs and triplane latent representation \cite{rodin,triplanediff} has gained significant popularity in recent years. In fact, after the announcement of the winners of the ICCV 2023 OmniObject3D Challenge, we found that almost all the competitors used DM-based approaches. However, as introduced in Section \ref{related} and Section \ref{mov}, the superiority of DMs lies in their stability during training and control during the generation process (\textit{i.e.}, conditional generation), while the OmniObject3D Challenge is an unconditional generation challenge. Thus, it is an interesting topic to explore how well 3D GANs can perform, especially when researchers tend to favor DMs as the preferred solution. From this perspective, our work provides some valuable and inspiring results, demonstrating that 3D GANs are still in the game.

\section{Conclusion}
In this paper, we present a solution for 3D object generation in the ICCV 2023 OmniObject3D Challenge. The key insight of our approach is to learn effective Neural Radiance Fields (NeRFs) and Signed Distance Functions (SDFs) representations using 3D generative adversarial networks (GANs), enabling high-fidelity 3D object synthesis from multiple viewpoints. Our experimental results demonstrate that the proposed method effectively generates high-quality images with fine details and accurate geometry. We further find that incorporating label embeddings is highly beneficial for improving the model’s learning effectiveness, leading to better generalization and performance. Additionally, our approach achieves competitive results compared to a state-of-the-art method proposed at CVPR 2023, further validating its effectiveness. Overall, our solution ranks among the top three in the ICCV 2023 OmniObject3D Challenge, highlighting that the 3D GAN-based method is competitive in an era where diffusion models are becoming the dominant approach for 3D generation.
\bibliographystyle{IEEEtran}
\bibliography{citations}

\begin{thebibliography}{10}
\providecommand{\url}[1]{#1}
\csname url@samestyle\endcsname
\providecommand{\newblock}{\relax}
\providecommand{\bibinfo}[2]{#2}
\providecommand{\BIBentrySTDinterwordspacing}{\spaceskip=0pt\relax}
\providecommand{\BIBentryALTinterwordstretchfactor}{4}
\providecommand{\BIBentryALTinterwordspacing}{\spaceskip=\fontdimen2\font plus
\BIBentryALTinterwordstretchfactor\fontdimen3\font minus \fontdimen4\font\relax}
\providecommand{\BIBforeignlanguage}[2]{{%
\expandafter\ifx\csname l@#1\endcsname\relax
\typeout{** WARNING: IEEEtran.bst: No hyphenation pattern has been}%
\typeout{** loaded for the language `#1'. Using the pattern for}%
\typeout{** the default language instead.}%
\else
\language=\csname l@#1\endcsname
\fi
#2}}
\providecommand{\BIBdecl}{\relax}
\BIBdecl

\bibitem{omni}
T.~Wu, J.~Zhang, X.~Fu, Y.~Wang, J.~Ren, L.~Pan, W.~Wu, L.~Yang, J.~Wang, C.~Qian \emph{et~al.}, ``Omniobject3d: Large-vocabulary 3d object dataset for realistic perception, reconstruction and generation,'' in \emph{Proc. of the IEEE/CVF Conference on Computer Vision and Pattern Recognition}, 2023, pp. 803--814.

\bibitem{jiang2023sdf}
L.~Jiang, R.~Ji, and L.~Zhang, ``Sdf-3dgan: A 3d object generative method based on implicit signed distance function,'' \emph{arXiv preprint arXiv:2303.06821}, 2023.

\bibitem{karras2019style}
T.~Karras, S.~Laine, and T.~Aila, ``A style-based generator architecture for generative adversarial networks,'' in \emph{Proceedings of the IEEE/CVF conference on computer vision and pattern recognition}, 2019, pp. 4401--4410.

\bibitem{esser2021taming}
P.~Esser, R.~Rombach, and B.~Ommer, ``Taming transformers for high-resolution image synthesis,'' in \emph{Proceedings of the IEEE/CVF conference on computer vision and pattern recognition}, 2021, pp. 12\,873--12\,883.

\bibitem{lunz2020inverse}
S.~Lunz, Y.~Li, A.~Fitzgibbon, and N.~Kushman, ``Inverse graphics gan: Learning to generate 3d shapes from unstructured 2d data,'' \emph{arXiv preprint arXiv:2002.12674}, 2020.

\bibitem{gadelha20173d}
M.~Gadelha, S.~Maji, and R.~Wang, ``3d shape induction from 2d views of multiple objects,'' in \emph{2017 International Conference on 3D Vision (3DV)}.\hskip 1em plus 0.5em minus 0.4em\relax IEEE, 2017, pp. 402--411.

\bibitem{nfi}
D.~Pavllo, D.~J. Tan, M.-J. Rakotosaona, and F.~Tombari, ``Shape, pose, and appearance from a single image via bootstrapped radiance field inversion,'' in \emph{Proc. of the IEEE/CVF Conference on Computer Vision and Pattern Recognition}, 2023, pp. 4391--4401.

\bibitem{mo2019structurenet}
K.~Mo, P.~Guerrero, L.~Yi, H.~Su, P.~Wonka, N.~Mitra, and L.~J. Guibas, ``Structurenet: Hierarchical graph networks for 3d shape generation,'' \emph{arXiv preprint arXiv:1908.00575}, 2019.

\bibitem{stylegan}
T.~Karras, S.~Laine, M.~Aittala, J.~Hellsten, J.~Lehtinen, and T.~Aila, ``Analyzing and improving the image quality of stylegan,'' in \emph{Proc. of the IEEE/CVF conference on computer vision and pattern recognition}, 2020, pp. 8110--8119.

\bibitem{eg3d}
E.~R. Chan, C.~Z. Lin, M.~A. Chan, K.~Nagano, B.~Pan, S.~De~Mello, O.~Gallo, L.~J. Guibas, J.~Tremblay, S.~Khamis \emph{et~al.}, ``Efficient geometry-aware 3d generative adversarial networks,'' in \emph{Proc. of the IEEE/CVF Conference on Computer Vision and Pattern Recognition}, 2022, pp. 16\,123--16\,133.

\bibitem{rodin}
T.~Wang, B.~Zhang, T.~Zhang, S.~Gu, J.~Bao, T.~Baltrusaitis, J.~Shen, D.~Chen, F.~Wen, Q.~Chen \emph{et~al.}, ``Rodin: A generative model for sculpting 3d digital avatars using diffusion,'' in \emph{Proc. of the IEEE/CVF Conference on Computer Vision and Pattern Recognition}, 2023, pp. 4563--4573.

\bibitem{txtmesh}
J.~Choi, D.~Jung, T.~Lee, S.~Kim, Y.~Jung, D.~Manocha, and D.~Lee, ``Tmo: Textured mesh acquisition of objects with a mobile device by using differentiable rendering,'' in \emph{Proc. of the IEEE/CVF Conference on Computer Vision and Pattern Recognition}, 2023, pp. 16\,674--16\,684.

\bibitem{nerf}
B.~Mildenhall, P.~P. Srinivasan, M.~Tancik, J.~T. Barron, R.~Ramamoorthi, and R.~Ng, ``Nerf: Representing scenes as neural radiance fields for view synthesis,'' \emph{Communications of the ACM}, vol.~65, no.~1, pp. 99--106, 2021.

\bibitem{pointnet}
C.~R. Qi, H.~Su, K.~Mo, and L.~J. Guibas, ``Pointnet: Deep learning on point sets for 3d classification and segmentation,'' in \emph{Proc. of the IEEE conference on computer vision and pattern recognition}, 2017, pp. 652--660.

\bibitem{mesh1}
H.~Ben-Hamu, H.~Maron, I.~Kezurer, G.~Avineri, and Y.~Lipman, ``Multi-chart generative surface modeling,'' \emph{ACM Transactions on Graphics}, vol.~37, no.~6, pp. 1--15, 2018.

\bibitem{voxel1}
X.~Wang, M.~H. Ang, and G.~H. Lee, ``Voxel-based network for shape completion by leveraging edge generation,'' in \emph{Proc. of the IEEE/CVF international conference on computer vision}, 2021, pp. 13\,189--13\,198.

\bibitem{deepsdf}
J.~J. Park, P.~Florence, J.~Straub, R.~Newcombe, and S.~Lovegrove, ``Deepsdf: Learning continuous signed distance functions for shape representation,'' in \emph{Proc. of the IEEE/CVF conference on computer vision and pattern recognition}, 2019, pp. 165--174.

\bibitem{mvdeepsdf}
Y.~Liu, K.~Zhu, G.~Wu, Y.~Ren, B.~Liu, Y.~Liu, and S.~Jinjun, ``Mv-deepsdf: Implicit modeling with multi-sweep point clouds for 3d vehicle reconstruction in autonomous driving,'' in \emph{Proc. of the IEEE/CVF International Conference on Computer Vision}, 2023.

\bibitem{occ}
L.~Mescheder, M.~Oechsle, M.~Niemeyer, S.~Nowozin, and A.~Geiger, ``Occupancy networks: Learning 3d reconstruction in function space,'' in \emph{Proc. of the IEEE/CVF conference on computer vision and pattern recognition}, 2019, pp. 4460--4470.

\bibitem{diffrf}
N.~M{\"u}ller, Y.~Siddiqui, L.~Porzi, S.~R. Bulo, P.~Kontschieder, and M.~Nie{\ss}ner, ``Diffrf: Rendering-guided 3d radiance field diffusion,'' in \emph{Proc. of the IEEE/CVF Conference on Computer Vision and Pattern Recognition}, 2023, pp. 4328--4338.

\bibitem{vae}
M.~Petrovich, M.~J. Black, and G.~Varol, ``Action-conditioned 3d human motion synthesis with transformer vae,'' in \emph{Proc. of the IEEE/CVF International Conference on Computer Vision}, 2021, pp. 10\,985--10\,995.

\bibitem{flow}
X.~Han, X.~Hu, W.~Huang, and M.~R. Scott, ``Clothflow: A flow-based model for clothed person generation,'' in \emph{Proc. of the IEEE/CVF international conference on computer vision}, 2019, pp. 10\,471--10\,480.

\bibitem{obja}
M.~Deitke, R.~Liu, M.~Wallingford, H.~Ngo, O.~Michel, A.~Kusupati, A.~Fan, C.~Laforte, V.~Voleti, S.~Y. Gadre \emph{et~al.}, ``Objaverse-xl: A universe of 10m+ 3d objects,'' \emph{Advances in Neural Information Processing Systems}, vol.~36, 2024.

\bibitem{shapenet}
A.~X. Chang, T.~Funkhouser, L.~Guibas, P.~Hanrahan, Q.~Huang, Z.~Li, S.~Savarese, M.~Savva, S.~Song, H.~Su \emph{et~al.}, ``Shapenet: An information-rich 3d model repository,'' \emph{arXiv preprint arXiv:1512.03012}, 2015.

\bibitem{ren}
Y.~Ren, G.~Wu, R.~Li, Z.~Yang, Y.~Liu, X.~Chen, T.~Cao, and B.~Liu, ``Unigaussian: Driving scene reconstruction from multiple camera models via unified gaussian representations,'' \emph{arXiv preprint arXiv:2411.15355}, 2024.

\bibitem{lpr}
Y.~Liu, J.~Shan, A.~Haridevan, and S.~Zhang, ``L-pr: Exploiting lidar fiducial marker for unordered low overlap multiview point cloud registration,'' \emph{arXiv preprint arXiv:2406.03298}, 2024.

\bibitem{ap3}
M.~Krogius, A.~Haggenmiller, and E.~Olson, ``Flexible layouts for fiducial tags,'' in \emph{Proc. IEEE/RSJ International Conference on Intelligent Robots and Systems}, 2019, pp. 1898--1903.

\bibitem{intensity}
Y.~Liu, H.~Schofield, and J.~Shan, ``Intensity image-based lidar fiducial marker system,'' \emph{IEEE Robotics and Automation Letters}, vol.~7, no.~3, pp. 6542--6549, 2022.

\bibitem{improvements}
Y.~Liu, J.~Shan, and H.~Schofield, ``Improvements to thin-sheet 3d lidar fiducial tag localization,'' \emph{IEEE Access}, 2024.

\bibitem{shuo1}
S.~Zhang, J.~Shan, and Y.~Liu, ``Variational bayesian estimator for mobile robot localization with unknown noise covariance,'' \emph{IEEE/ASME Transactions on Mechatronics}, vol.~27, no.~4, pp. 2185--2193, 2022.

\bibitem{liao}
T.~Liao, A.~Haridevan, Y.~Liu, and J.~Shan, ``Autonomous vision-based uav landing with collision avoidance using deep learning,'' in \emph{Science and Information Conference}.\hskip 1em plus 0.5em minus 0.4em\relax Springer, 2022, pp. 79--87.

\bibitem{shuo2}
S.~Zhang, J.~Shan, and Y.~Liu, ``Approximate inference particle filtering for mobile robot slam,'' \emph{IEEE Transactions on Automation Science and Engineering}, 2024.

\bibitem{navigation}
Y.~Liu, H.~Schofield, and J.~Shan, ``Navigation of a self-driving vehicle using one fiducial marker,'' in \emph{2021 IEEE International Conference on Multisensor Fusion and Integration for Intelligent Systems (MFI)}.\hskip 1em plus 0.5em minus 0.4em\relax IEEE, 2021, pp. 1--6.

\bibitem{shuo3}
S.~Zhang, J.~Shan, and Y.~Liu, ``Particle filtering on lie group for mobile robot localization with range-bearing measurements,'' \emph{IEEE Control Systems Letters}, 2023.

\bibitem{deblur}
Y.~Liu, A.~Haridevan, H.~Schofield, and J.~Shan, ``Application of ghost-deblurgan to fiducial marker detection,'' in \emph{2022 IEEE/RSJ International Conference on Intelligent Robots and Systems (IROS)}.\hskip 1em plus 0.5em minus 0.4em\relax IEEE, 2022, pp. 6827--6832.

\bibitem{vqadiff}
Y.~Liu, Z.~Yang, G.~Wu, Y.~Ren, K.~Lin, B.~Liu, Y.~Liu, and J.~Shan, ``Vqa-diff: Exploiting vqa and diffusion for zero-shot image-to-3d vehicle asset generation in autonomous driving,'' in \emph{European Conference on Computer Vision}.\hskip 1em plus 0.5em minus 0.4em\relax Springer, 2024, pp. 323--340.

\bibitem{neus}
P.~Wang, L.~Liu, Y.~Liu, C.~Theobalt, T.~Komura, and W.~Wang, ``Neus: Learning neural implicit surfaces by volume rendering for multi-view reconstruction,'' \emph{arXiv preprint arXiv:2106.10689}, 2021.

\bibitem{gan2}
A.~Trevithick, M.~Chan, T.~Takikawa, U.~Iqbal, S.~De~Mello, M.~Chandraker, R.~Ramamoorthi, and K.~Nagano, ``What you see is what you gan: Rendering every pixel for high-fidelity geometry in 3d gans,'' in \emph{Proceedings of the IEEE/CVF Conference on Computer Vision and Pattern Recognition}, 2024, pp. 22\,765--22\,775.

\bibitem{triplanediff}
J.~R. Shue, E.~R. Chan, R.~Po, Z.~Ankner, J.~Wu, and G.~Wetzstein, ``3d neural field generation using triplane diffusion,'' in \emph{Proceedings of the IEEE/CVF Conference on Computer Vision and Pattern Recognition}, 2023, pp. 20\,875--20\,886.

\bibitem{ada}
T.~Karras, M.~Aittala, J.~Hellsten, S.~Laine, J.~Lehtinen, and T.~Aila, ``Training generative adversarial networks with limited data,'' \emph{Advances in neural information processing systems}, vol.~33, pp. 12\,104--12\,114, 2020.

\end{thebibliography}
\end{document}